\pdfoutput=1

\documentclass[11pt]{article}

\usepackage[preprint]{acl}

\usepackage{times}
\usepackage{latexsym}
\usepackage{hyperref}

\usepackage[T1]{fontenc}

\usepackage[utf8]{inputenc}

\usepackage{microtype}

\usepackage{inconsolata}

\usepackage{graphicx}
\usepackage{amsmath}

\usepackage{hyperref}
\usepackage{url}

\usepackage{graphicx}
\usepackage{wrapfig}
\usepackage{tikz}
\usepackage{subfigure}
\usepackage{float}
\usepackage{caption}
\usepackage{enumerate}
\usepackage{array}
\usepackage[ruled, linesnumbered]{algorithm2e}
\usepackage{hyperref}
\usepackage{url}
\usepackage{todonotes}
\usepackage{ulem}

\usepackage{colortbl}

\usepackage{rotating}
\usepackage{booktabs}
\usepackage{multirow}
\usepackage{makecell}
\usepackage{amssymb}

\usepackage{amsmath}

\usepackage{amssymb}
\usepackage{pdfrender}

\usepackage{xcolor}

\SetKwInput{KwInput}{Input}                
\SetKwInput{KwOutput}{Output}              
\SetKwInput{KwData}{Data}

\title{How Privacy-Savvy Are Large Language Models? A Case Study on \\
Compliance and Privacy Technical Review}

\author{Yang Liu${}^*$, Xichou Zhu${}^*$\\
\textbf{Zhou Shen, Yi Liu, Min Li, Yujun Chen,
Benzi John, Zhenzhen Ma, Tao Hu, Zhi Li,}\\
\textbf{Bolong Yang, Manman Wang, Zongxing Xie, Peng Liu, Dan Cai, Junhui Wang}\\
  Machine Learning \& AI Team, Privacy and Data Protection Office\\
  ByteDance, Beijing, China\\
  * denotes equal contribution \\
  \texttt{\{zhuxichou, liuyang.173, junhui.wang\}@bytedance.com} 
}

\begin{document}
\maketitle
\begin{abstract}

The recent advances in large language models (LLMs) have significantly expanded their applications across various fields such as language generation, summarization, and complex question answering. However, their application to privacy compliance and technical privacy reviews remains under-explored, raising critical concerns about their ability to adhere to global privacy standards and protect sensitive user data. This paper seeks to address this gap by providing a comprehensive case study evaluating LLMs' performance in privacy-related tasks such as privacy information extraction (PIE), legal and regulatory key point detection (KPD), and question answering (QA) with respect to privacy policies and data protection regulations. We introduce a Privacy Technical Review (PTR) framework, highlighting its role in mitigating privacy risks during the software development life-cycle. Through an empirical assessment, we investigate the capacity of several prominent LLMs, including BERT, GPT-3.5, GPT-4, and custom models, in executing privacy compliance checks and technical privacy reviews. Our experiments benchmark the models across multiple dimensions, focusing on their precision, recall, and F1-scores in extracting privacy-sensitive information and detecting key regulatory compliance points. While LLMs show promise in automating privacy reviews and identifying regulatory discrepancies, significant gaps persist in their ability to fully comply with evolving legal standards. We provide actionable recommendations for enhancing LLMs' capabilities in privacy compliance, emphasizing the need for robust model improvements and better integration with legal and regulatory requirements. This study underscores the growing importance of developing privacy-aware LLMs that can both support businesses in compliance efforts and safeguard user privacy rights.
\end{abstract}

\section{Introduction}

Large language models (LLMs) \citep{yao2024survey} have become increasingly influential in various sectors due to their proficiency in tasks such as language generation, summarization, and question answering. With advancements in natural language processing, LLMs are now being integrated into a wide range of applications \citep{topsakal2023creating, liu2023prompt}. However, their application in the domain of privacy compliance and technical privacy reviews has not been thoroughly examined. 

Also, the increasing reliance on LLMs in various applications has raised important questions about their privacy awareness \citep{yao2024survey}. As these models are deployed across sensitive industries \citep{pankajakshan2024mapping}—such as healthcare \citep{gebreab2024llm}, finance \citep{zhao2024revolutionizing}, and legal services \citep{sun2024lawluo}—ensuring their compliance with privacy regulations. Given the growing concerns about data privacy and security, ensuring that LLMs can effectively handle privacy-sensitive tasks is crucial. Regulatory frameworks such as the General Data Protection Regulation and the California Consumer Privacy Act have set stringent requirements for data protection, and as these models are applied to sensitive domains, their ability to comply with such regulations must be scrutinized.

\begin{figure*}[t]
    \centering
    \includegraphics[width=1.0\linewidth]{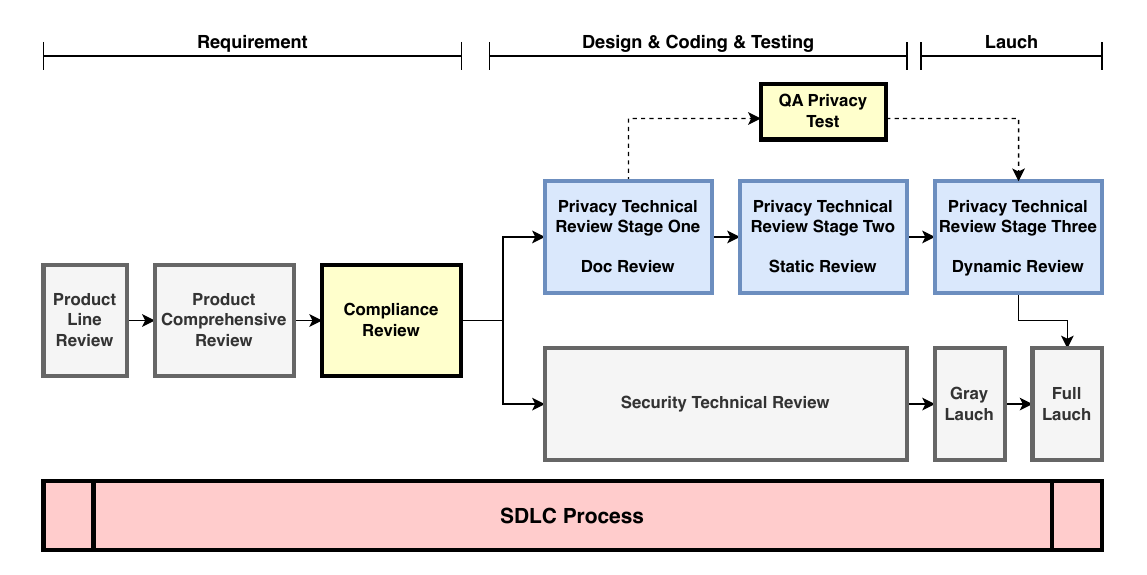}
    \caption{Illustration of software development life-cycle (SDLC) process.}
    \label{fig:sdlc}
\end{figure*}

This paper aims to bridge this gap by conducting a case study that evaluates the performance of LLMs in privacy-related tasks. Specifically, we assess how well these models perform in entity extraction, text classification, and question answering within the context of privacy compliance checks and technical privacy reviews. These tasks are integral to understanding whether LLMs can recognize and adhere to complex legal and technical privacy standards.

By investigating the strengths and limitations of LLMs in these specific areas, we hope to provide a clearer picture of their current capabilities in privacy compliance. Furthermore, our study will offer recommendations for improving LLMs to better align with the evolving demands of privacy regulations and technical requirements in the future.

Our contributions are summarized as follows:

\begin{itemize}
    \item \textbf{Introduction of Privacy Technical Review (PTR) Process.} We outline a detailed workflow for Privacy Technical Review (PTR), emphasizing its importance in identifying and mitigating privacy risks during the software development lifecycle. The PTR framework is illustrated as a comprehensive method to ensure compliance with privacy regulations.
    \item \textbf{Evaluation of LLMs in Privacy Compliance and Technical Privacy Reviews.} We present an empirical evaluation of LLMs in the context of privacy compliance and technical privacy reviews. The study specifically focuses on tasks such as privacy information extraction (PIE), key point detection (KPD) related to legal and regulatory issues, and question answering (QA) regarding privacy policies.
    \item \textbf{Experimental Benchmarking of Models.} The study benchmarks various LLMs, including BERT, GPT, and custom models, across the PIE, KPD, and QA tasks, providing insights into their precision, recall, and F1 scores, and highlighting the strengths and weaknesses of each model in privacy-related tasks.
\end{itemize}

\section{Related Works}

\paragraph{LLMs development} The development of Large Language Models (LLMs) marks a pivotal advancement in artificial intelligence \cite{zhao2023survey, yao2024survey, chen2024entity, zhang2025surveygraphretrievalaugmentedgeneration}, particularly in the domain of natural language processing. Building on deep learning techniques, especially transformer architectures like GPT \cite{radford2018improving} and BERT \cite{devlin-etal-2019-bert}, LLMs have leveraged massive datasets and billions of parameters to achieve unprecedented capabilities in language understanding \cite{nam2024using} and generation \cite{chang2023learning}. These models excel in tasks ranging from text generation \cite{mo2024large} and translation to code creation and summarization \cite{jin2024comprehensive, laban2023summedits}, and their scalability has enabled them to demonstrate remarkable proficiency across diverse applications. However, as LLMs continue to expand in capability, ethical concerns such as bias \cite{lin2024investigating} and privacy issues \cite{zhang2024no, yao2024survey} have come to the forefront, prompting a concerted effort toward more responsible AI development. This evolving landscape underscores the transformative potential of LLMs in both digital and real-world contexts while emphasizing the need for ongoing research in their ethical deployment.

\paragraph{LLMs and Privacy} Recent advancements in Large Language Models (LLMs) have spurred significant interest in their privacy and legal implications \cite{yao2024survey, trautmann2022legal}. Previous works have explored the vulnerabilities \cite{fang2024llm} of LLMs to privacy risks \cite{wu2024new}, such as data leakage \cite{wang2024pandora, borkar2023can}, model inversion attacks \cite{chen2024text}, and membership inference \cite{galli2024noisy, feng2024exposing, kaneko2024sampling}, which can expose sensitive information. To address these concerns, researchers have focused on improving model architectures, applying differential privacy techniques, and implementing access control mechanisms. Moreover, legal frameworks \cite{raucea2019llm, patronidi2021implications} such as the GDPR and CCPA have prompted studies on the compliance of LLMs with data protection regulations. These works highlight the growing importance of integrating technical solutions with legal obligations to ensure that LLMs not only perform effectively but also respect user privacy and align with regulatory standards.

\section{Background}

\begin{figure}[t]
    \centering
    \includegraphics[width=1.0\linewidth]{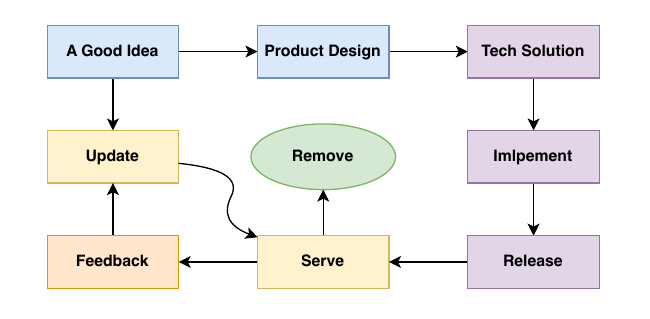}
    \caption{Illustration of privacy technical review in business process.}
    \label{fig:ptr_bp}
\end{figure}

\subsection{Compliance}

Data compliance refers to the adherence to laws, regulations, and guidelines that govern the collection, storage, processing, and sharing of data. It involves ensuring that an organization's data handling practices align with legal requirements such as data protection laws (e.g., GDPR, CCPA) and industry standards. The goal of data compliance is to protect personal and sensitive information, ensure privacy, and mitigate risks associated with data breaches and misuse. Compliance requires organizations to implement proper policies, procedures, and technical measures to manage data responsibly and ethically.

\subsection{Privacy Technical Review}

Privacy Technical Review (PTR) is a systematic technical assessment process aimed at identifying privacy risks that a company may have failed to effectively address during the design and implementation of its products or services. This evaluation is based on legal requirements and legal compliance assessments from a technical perspective. PTR is conducted to verify whether a product complies with relevant privacy regulations before and shortly after its launch. By performing PTR, product development teams can be guided to correctly understand and follow Privacy by Design principles, comprehend the privacy implications of their data processing activities, and identify and mitigate potential privacy risks. Ultimately, PTR enhances the level of user privacy protection and safeguards user privacy rights.

PTR aims to identify and mitigate potential privacy risks in products and systems before they go live. It helps business teams discover these risks and resolve them within a closed-loop system prior to product launch. The core activities of PTR include reviewing requirement documents, technical solutions, code, and traffic. These reviews are based on global privacy and data protection laws, privacy design principles, relevant company systems, and legal opinions. PTR is a key method for implementing PDPS (Personal Data Protection System) and PLR (Privacy Legal Review). Through cross-functional collaboration, PTR ensures the effective enforcement of privacy and data protection requirements while enhancing business compliance.

The significance of involving large models in PTR lies in their ability to efficiently handle complex and large-scale privacy review tasks. With their advanced natural language processing capabilities, large models can assist in identifying policies, specifications, and privacy risks, thus supporting business decision-making. These models can propose technical solutions based on upstream legal and business requirements, and assess whether these solutions effectively reduce privacy risks. Furthermore, through automation and intelligent processing, large models can alleviate the burden of PTR, reducing complexity and time consumption, making the privacy review process more efficient for businesses.

\begin{figure}[t]
    \centering
    \includegraphics[width=1.0\linewidth]{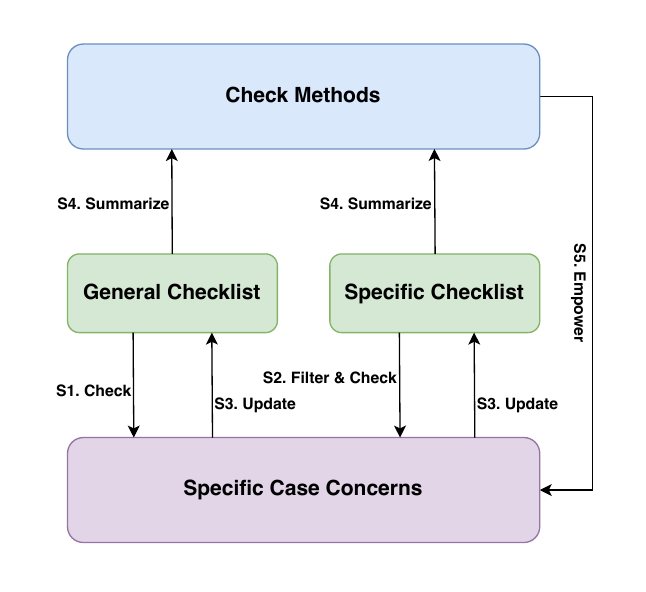}
    \caption{Illustration of privacy technical review process.}
    \label{fig:ptrp}
\end{figure}

\begin{figure*}[t]
    \centering
    \includegraphics[width=.9\linewidth]{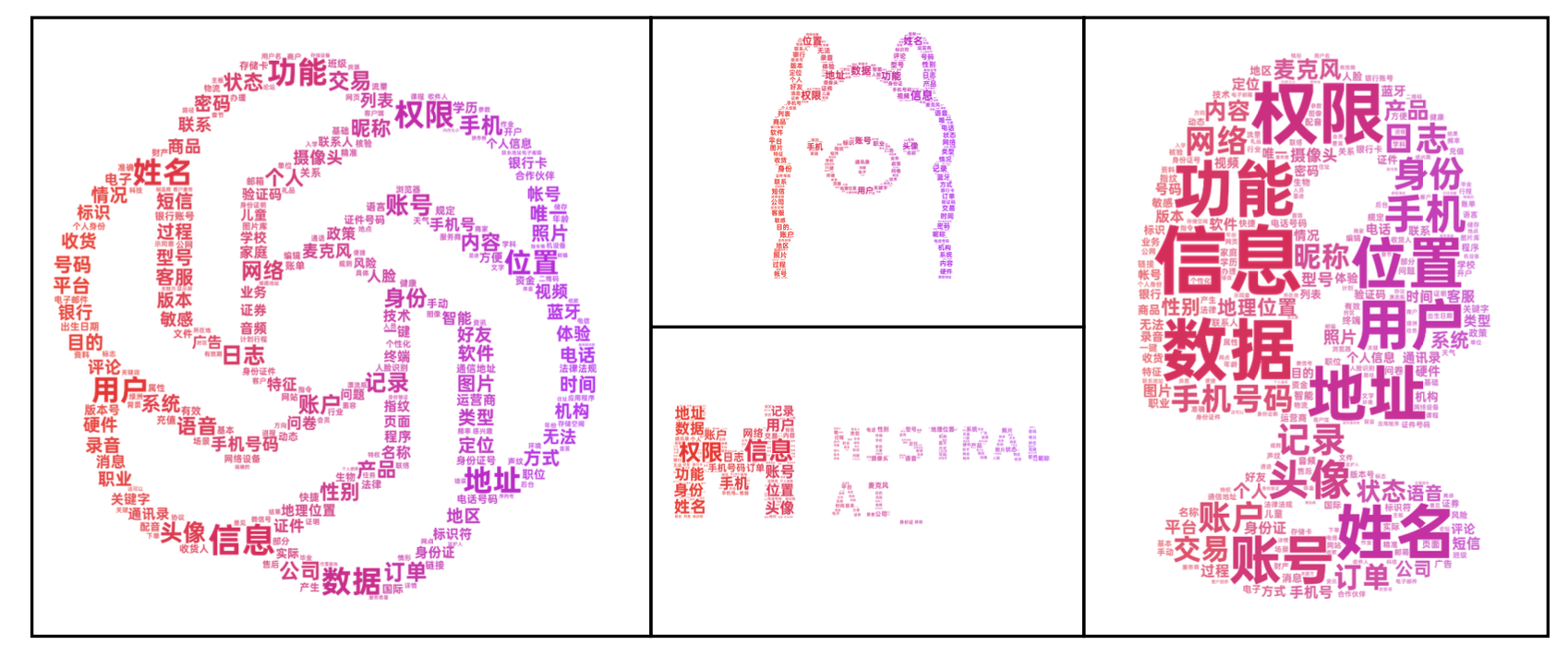}
    \caption{Word cloud of QA dataset.}
    \label{fig:wordcloud}
\end{figure*}

\subsection{PTR in Business}

In business, the Privacy Technical Review (PTR) plays a critical role in ensuring privacy compliance throughout the software development lifecycle (SDLC). As seen in the workflow diagram, PTR has expanded from the design phase to all stages of the SDLC, encompassing initial product design, technical solutions, implementation, and eventual updates or removal. This comprehensive approach allows for continuous privacy assessments, from conception to decommissioning, ensuring that all changes, updates, and optimizations are thoroughly evaluated for privacy risks. The PTR iteration process further highlights its significance, as it inputs key artifacts such as PRDs, ERDs, and legal tickets. Although the product may not be fully developed, PTR focuses on detecting potential deficiencies or risks in technical solutions that could compromise privacy. By following check methods that include general and specific checklists, the process filters, checks, updates, and summarizes findings to empower the team to address specific privacy concerns in a structured and methodical manner.

\section{Evaluation Tasks}

\paragraph{Privacy Information Extraction (PIE)} This task focuses on extracting privacy-related information from protocol texts. The open-source test set consists of approximately 8,800 sentences. The annotation follows the BIOE labeling scheme, which is commonly used in Named Entity Recognition (NER) tasks. The goal is to identify the types of privacy data collected as declared in the text. The tag-to-index mapping is defined as {O:0, B:1, I:3, E:2}.

\paragraph{Key Point Detection in Legal and Regulatory (KPD)} The objective of this task is to detect key legal regulatory points. The dataset contains 10 key legal regulatory aspects, similar in scale to the Privacy Information Extraction dataset. These key points include:
\begin{enumerate}
    \item ExceedLimit: Handling of personal information beyond its retention period.
    \item StorageRegion: Information on the location of personal data storage.
    \item StorageTime: Duration for which personal information is stored.
    \item Aging: Timeliness of the privacy policy.
    \item Query: Description of the right to query personal information (User Rights Protection).
    \item Correct: Description of the right to correct personal information (User Rights Protection).
    \item Delete: Description of the right to delete personal information (User Rights Protection).
    \item Logout: Disposal of personal information upon account deletion (User Rights Protection).
    \item SDK: Information on the usage of SDKs.
    \item Repeal: Explanation of the method to revoke authorization (User Rights Protection).
\end{enumerate}

\paragraph{Question and Answer (QA)} This task evaluates reading comprehension ability and follows the format of Extractive Machine Reading Comprehension (MRC) datasets. Given a combination of a protocol text and a query, the model must locate the appropriate answer within the protocol text. This dataset comprises approximately 2,300 protocol texts, each associated with one query. The key characteristic of this dataset is that the answer appears only once in the text and occupies a single continuous span of text.

\begin{table}[t]
\small
\centering
\caption{Results on PIE task (\%). 
}
\label{table:PIE}
\begin{tabular}{lccc}
\toprule
\textbf{Models} & \textbf{Precision} & \textbf{Recall} & \textbf{F1-score} \\
\midrule
BERT-base &92.4 &79.1 &85.0\\
roberta-wwm-ext-large &90.0 &87.1 &88.4\\
ComBERT &90.8 &92.9 &91.8\\
\midrule
GPT-3.5-turbo &98.4 &98.9 &98.4 \\
GPT-4 &98.8 &99.5 &99.0 \\
GPT-4o &99.6 &99.7 &99.6 \\
Mistral\_7b &97.6 &97.9 &97.7 \\
gemini-1.5-flash &99.8 &99.9 &99.8 \\
moonshot\_8k\_v1 &99.7 &99.7 &99.7 \\
Doubao &96.4 &97.7 &96.7 \\
Doubao-pro &96.1 &96.6 &96.2 \\
\bottomrule
\end{tabular}
\end{table}

\subsection{Evaluation Metrics}

\paragraph{PIE Metrics} The evaluation for PIE uses the BIOE Macro-Averaging Metrics. The labels (B, I, O, E) are evaluated separately, with precision, recall, and F1-score calculated for each label. The final scores are obtained by averaging the precision, recall, and F1-scores across the four labels, resulting in Macro-Averaging Precision, Recall, and F1-Score.

\paragraph{KPD Metrics} The evaluation for KPD involves calculating:

\begin{itemize}
    \item Precision. Precision is computed for each label individually, and then the average across all labels is taken.
    \item Recall. Recall is computed for each label individually, and then the average across all labels is taken.
    \item F1-Score. The F1-score is calculated for each label individually, and then the average across all labels is taken.
\end{itemize}

\paragraph{QA Metrics} The evaluation for QA involves
calculating:

\begin{itemize}
    \item ROUGE-L. This measures the fuzzy matching between the predicted answer and the ground truth answer at the word level. It does not treat the answers as "bags of words," but instead calculates the Longest Common Subsequence (LCS) between them. Based on this, precision (P), recall (R), and F1-score (F1) are computed.
    \item Exact Match (EM). This is a binary metric that gives a score of 1 if the predicted answer exactly matches the ground truth answer, and 0 otherwise.
    \item Re-85. This is a fuzzy matching metric where a prediction is considered correct if the fuzzy matching score is greater than 85.
\end{itemize}

These metrics provide a comprehensive evaluation of each task by considering different aspects of accuracy, recall, and fuzzy matching.

\section{Experiments}

\begin{table}[t]
\small
\centering
\caption{Results on KPD task (\%). 
}
\label{table:KPD}
\begin{tabular}{lccc}
\toprule
\textbf{Models} & \textbf{Precision} & \textbf{Recall} & \textbf{F1-score} \\
\midrule
BERT-base &87.8 &94.3 &90.7\\
roberta-wwm-ext-large &92.4 &92.5 &91.8\\
ComBERT &94.2 &94.1 &94.0\\
\midrule
GPT-3.5-turbo &95.0 &94.5 &94.4\\
GPT-4 &93.1 &92.5 &94.8\\
GPT-4o &96.0 &94.4 &94.5\\
Mistral\_7b &88.1 &88.4 &93.5\\
gemini-1.5-flash &95.4 &95.9 &94.4\\
moonshot\_8k\_v1 &92.3 &94.0 &94.2\\
Doubao &90.0 &86.7 &88.1\\
Doubao-pro &94.4 &95.2 &92.0\\
\bottomrule
\end{tabular}
\end{table}

\begin{table*}[t]
\small
\centering
\caption{Results on QA task (\%).
}
\label{table:QA}
\begin{tabular}{lccccc}
\toprule
\textbf{Models} & \textbf{Precision} & \textbf{Recall} & \textbf{F1-score} & \textbf{EM} & \textbf{RE\_85}\\
\midrule
alicemind &5.1	&5.4 &5.0 &1.6 &7.4\\
SiameseUIE &17.6 &17.8 &15.9 &9.9 &26.1\\
erlangshen &57.5 &48.0 &48.0 &31.0 &76.4\\
ComBERT &81.8 &60.9 &67.5 &54.9 &96.9\\
\midrule
GPT-3.5-turbo &64.5 &100.0 &67.9 &40.0 &81.1\\
GPT-4 &58.8 &100.0 &72.8 &40.5 &33.3\\
GPT-4o &67.6 &100.0 &70.6 &34.0 &63.3 \\
Mistral\_7b &55.4 &95.4 &68.3 &34.0 &9.6\\
gemini-1.5-flash &49.8 &100.0 &62.0 &33.0 &56.7\\
moonshot\_8k\_v1 &52.3 &100.0 &66.0 &53.2 &36.6\\
Doubao &62.8 &97.8 &95.4 & 32.0 &78.4\\
Doubao-pro &68.8 &100.0 &73.6 & 34.5 &78.1\\
\bottomrule
\end{tabular}
\end{table*}

\subsection{Dataset}

The datasets\footnote{\url{https://github.com/alipay/ComBERT}} used in this study encompass a range of text extraction and comprehension tasks across different domains. Here we give a brief introduction for these datasets:
\begin{enumerate}
    \item Privacy Information Extraction Dataset. This dataset comprises approximately 8,800 sentences extracted from privacy policies and agreement texts. Each sentence is annotated using a BIOE tagging scheme to identify personal data types, such as names, gender, ID numbers, and transaction information. The annotations provide detailed information regarding the specific personal data mentioned in the text.
    \item Legal and Regulatory Key Point Detection Dataset. This dataset focuses on identifying key legal and regulatory requirements within text. It defines 10 key legal concepts, including data retention periods, methods for deleting personal data, and the validity of privacy policies. Each sentence in the dataset is labeled with a binary value indicating whether it contains one or more of these key legal points.
    \item Domain-Specific Question Answering Dataset. This dataset contains roughly 2,300 passages from agreement texts, each paired with a query. The task is to locate a contiguous span of text within the passage that directly answers the query. The dataset is well-suited for evaluating models in domain-specific question answering and information retrieval scenarios.
\end{enumerate}

These datasets provide a comprehensive foundation for evaluating text extraction and comprehension capabilities across diverse legal and privacy-related domains.

\subsection{Baselines}

Here we choose several LLMs as the baselines including GPT-3.5-turbo, GPT-4, GPT-4o\footnote{\url{https://platform.openai.com/docs/models}}, Mistral\_7b\footnote{\url{https://mistral.ai/news/announcing-mistral-7b/}}, gemini-1.5-flash\footnote{\url{https://deepmind.google/technologies/gemini/flash/}}, moonshot\_8k\_v1\footnote{\url{https://platform.moonshot.cn/}}, doubao and doubao-pro\footnote{\url{https://www.volcengine.com/}}.

\subsection{Results and Discussions}

Table \ref{table:PIE} shows the performance of various models on PIE. In the PIE task, large language models, particularly those from the GPT series, demonstrate outstanding performance. For instance, GPT-3.5-turbo achieves an F1-score of 98.4, while GPT-4 and GPT-4o further improve this to 99.0 and 99.6, respectively. The gemini-1.5-flash model exhibits near-perfect performance with a Precision of 99.8 and Recall of 99.9, leading to an F1-score of 99.8. These results suggest that the latest generation of large models is capable of near-optimal accuracy across different metrics. In contrast, traditional baseline models such as BERT-base, roberta-wwm-ext-large, and ComBERT lag behind significantly, with the best-performing traditional model, ComBERT, reaching an F1-score of 91.8. This gap highlights the dramatic advancements made by large models in natural language processing tasks like PIE.

Table \ref{table:KPD} shows the performance of various models on KPD task. Similarly, in the KPD task, large language models again outperform traditional baselines by a significant margin. GPT-4o stands out with an F1-score of 94.5, slightly ahead of GPT-4, which scores 94.8 in this task. While the Mistral\_7b model shows a slight decline in performance compared to its counterparts, achieving an F1-score of 93.5, models like gemini-1.5-flash maintain high effectiveness with an F1-score of 94.4. It's noteworthy that even models like moonshot\_8k\_v1 and Doubao-pro manage to deliver competitive results, with F1-scores of 94.2 and 92.0, respectively. These findings further confirm the dominance of advanced models in complex NLP tasks, as they consistently surpass traditional models like BERT and ComBERT, which serve more as benchmarks rather than true competitors in these tasks.

Table \ref{table:QA} shows the performance of various models on a QA task. Overall, ComBERT stands out with strong performance across all metrics, particularly in F1-score (67.5) and RE\_85 (96.9), demonstrating good balance. Doubao and Doubao-pro lead in recall (100.0) and F1-score (95.4 and 73.6). In contrast, the GPT series models excel in recall but fall slightly behind in precision and EM. Pretrained model alicemind and SiameseUIE perform the worst, struggling to adapt to this task.

\section{Conclusion}

This paper provides a critical analysis of LLMs' performance in privacy-sensitive domains, highlighting both their potential and limitations in adhering to privacy standards. The findings suggest that while LLMs can assist in privacy reviews, further improvements are needed to fully align them with regulatory and technical privacy requirements. Recommendations for enhancing LLM capabilities are also provided, ensuring they can better meet the evolving demands of privacy compliance.

\bibliography{custom}




\end{document}